\pdfoutput=1
\documentclass[12pt]{article}

\usepackage{acl}

\usepackage{fancyhdr}
\usepackage{booktabs}
\usepackage[T1]{fontenc}
\usepackage{graphicx}
\graphicspath{ {./images/} }
\usepackage[utf8]{inputenc}
\usepackage{natbib}
\bibpunct{[}{]}{;}{s}

\title{Clickbait Classification and Spoiling Using Natural Language Processing}

\author{Adhitya Thirumala \and Elisa Ferracane}

\begin{document}
\maketitle
\begin{abstract}
Clickbait is the practice of engineering titles to incentivize readers to click through to articles. Such titles with sensationalized language reveal as little information as possible. Occasionally, clickbait will be intentionally misleading, so natural language processing (NLP) can scan the article and answer the question posed by the clickbait title, or spoil it. We tackle two tasks: classifying the clickbait into one of 3 types (Task 1), and spoiling the clickbait (Task 2). For Task 1, we propose two binary classifiers to determine the final spoiler type. For Task 2, we experiment with two  approaches: using a question-answering model to identify the span of text of the spoiler, and using a large language model (LLM) to generate the spoiler. Because the spoiler is contained in the article, we frame the second task as a question-answering approach for identifying the starting and ending positions of the spoiler. We created models for Task 1 that were better than the baselines proposed by the dataset authors and engineered prompts for Task 2 that did not perform as well as the baselines proposed by the dataset authors due to the evaluation metric performing worse when the output text is from a generative model as opposed to an extractive model.
\end{abstract}

\section{Introduction}

Clickbait is the practice of engineering titles to incentivize readers to click through to articles. It is an issue that plagues the internet, and spreads disinformation while causing readers to go through extreme methods to find the real content of an article. Many people, however, may not go to such an extent to find answers and will instead believe misleading information in the clickbait titles. Effectively, many may "judge a book by its cover." Such a problem can be solved by "adding to the cover" and creating a natural language processing model to "spoil" the article and reveal a short summary which answers the question posed by the clickbait title. We tackle two tasks that can help with the spoiling clickbait. Task 1 is classification, where the data is split into three categories to make the next task easier. Task 2 is spoiling, where we reveal the answer to the title. Task 1 makes Task 2 much easier because we can develop multiple algorithms to tackle the multiple types of clickbait. 

\begin{table}[t!]
\small
    \begin{center}
     \begin{tabular}{lp{4.5cm}}
\toprule
\texttt{targetTitle} & Google paid HOW MUCH in overseas taxes!? \\
\texttt{spoiler} & had a UK tax bill of 35 million pounds (\$55 million) in 2012 \\
\texttt{targetParagraphs} & Google, which has been grilled twice in the past year by a UK parliamentary committee over its tax practices, \textcolor{blue}{had a UK tax bill of 35 million pounds (\$55 million) in 2012}, on sales of \$4.9 billion to British customers, its accounts showed. The Internet search giant paid a tax rate of 2.6 percent on \$8.1 billion in non-U.S. income in 2012, because it channelled almost all of its overseas profits to a subsidiary in Bermuda which levies no corporate income tax [...]\\

\texttt{tags} & passage \\
\\ \bottomrule
\end{tabular}
\caption{Example of a clickbait (\texttt{targetTitle}), its \texttt{spoiler}, the full text of the linked article with the spoiler in blue (\texttt{targetParagraphs},  shortened for space) and the spoiler type (\texttt{tags}).}
\label{tab:data_example}
      \end{center}
      \end{table}
 
\section{Data}
We obtain our dataset from a shared task called SemEval. SemEval is a group and runs 12 tasks per year; we used the dataset from the Clickbait shared task. Specifically, the organizers of our tasks published a paper in which they defined a task, created a dataset, and made baseline classifiers. \cite{hagen_frobe_jurk_potthast_2022} This dataset contains 4000 entries: 3200 train and 800 validation. 17.4\% of the train dataset is \verb|multi|, 39.8\% of the train is \verb|passage|, and the remaining 42.8\% is \verb|phrase|. The dataset contains text and various metadata. Initial experiments found that the metadata did not help with predictions, so we do not utilize it in any predictions. The relevant parts of the dataset are the \texttt{targetParagraphs} column, the \texttt{targetTitle} column, the \texttt{spoiler} column, and the \texttt{tags} column. The \texttt{targetParagraphs} column contains the text of an article that contains the spoiler. In the dataset, it is formatted as a list where each element is a line of the article. The \texttt{targetTitle} contains the clickbait title of the article. The \texttt{spoiler} column contains the spoiler to the article. This spoiler is always a direct span of the text. The \texttt{tags} are the type of spoiler. There are three types of spoilers that we classify: \texttt{phrase}, \texttt{passage}, and \texttt{multi}. The \texttt{phrase} spoiler is always a span from the text less than or equal to 5 words in length. The \texttt{passage} spoiler is a single span of text longer than five words in length. The \texttt{multi} spoiler is multiple non-continuous spans of text. An example entry of the dataset's relevant columns is illustrated in Table \ref{tab:data_example}.

% \subsection{Example}
% \verb|targetTitle|: Google paid HOW MUCH in overseas taxes!? \\
% \verb|targetParagraphs|: Google, which has been grilled twice in the past year by a UK parliamentary committee over its tax practices, had a UK tax bill of 35 million pounds (\$55 million) in 2012, on sales of \$4.9 billion to British customers, its accounts showed. The Internet search giant paid a tax rate of 2.6 percent on \$8.1 billion in non-U.S. income in 2012, because it channelled almost all of its overseas profits to a subsidiary in Bermuda which levies no corporate income tax, the group's accounts show. \footnote{Article was truncated from actual dataset entry.}\\
% \verb|spoiler|: had a UK tax bill of 35 million pounds (\$55 million) in 2012 \\
% \verb|tags|: passage

% cite this ig
\section{Related Work}
Previously, many have taken on the task of detecting clickbait, such as approaches which not only detect the presence of clickbait, but also sort out and puts into categories based on their content such as inflammatory clickbait, ambiguous clickbait, and exaggeration clickbait. \cite{pujahari_sisodia_2019} This method allowed for detection, but it was incapable of actually solving the clickbait issue. There have also been attempts to generate clickbait using NLP. These approaches use a summarization method to make sensationalized headlines for articles. \cite{xu_wu_madotto_fung} This work makes the problem worse as it removes the human aspect of creating clickbait and makes the generation of clickbait easier and quicker.

The organizers of the shared task are the first to define the task of spoiling and classifying clickbait that we use, and also creates a dataset for the purpose. They released baseline models for both classification and spoiling; however, the spoiling models were not capable of handling entries that were tagged as \texttt{multi}. The baseline value that they obtain for Task 1 is 71.57\% balanced accuracy. For Task 2, they obtain a 0.688 for phrase spoiler and a 0.3144 for passage spoilers. These values are for the BLEU-4 score which will be discussed in section \ref{evalmetric}. Existing approaches to Task 2 take on an extractive question answering approach where the title of the article functions as the question and the text of the article functions as the context.
\section{Task 1: Clickbait Classification}
The first task that we tackle is classification of the clickbait into the three types shown above.
\subsection{Input/Output}
The inputs are the title of the article, and the output is one of the three types referenced above: \texttt{phrase}, \texttt{passage} and \texttt{multi}. We chose to only input the title because we found that the text of the article had little no no correlation with the performance of the model.
 \subsection{Approach}
We defined two models for this task. Both models are binary classifiers made using the deBERTa \cite{DBLP:journals/corr/abs-2006-03654}
neural network model for sequence classification. The first model that we use classifies between the \texttt{multi} spoilers and the other types of spoilers. We refer to this model as Model 1 in the section hereafter. The second model classifies the \texttt{passage} versus the \texttt{phrase} spoiler. We refer to this model as Model 2 in this section hereafter.
\subsection{Hyperparameter Optimization}
\begin{figure*}[t]
    \centering
    \includegraphics[width=15cm]{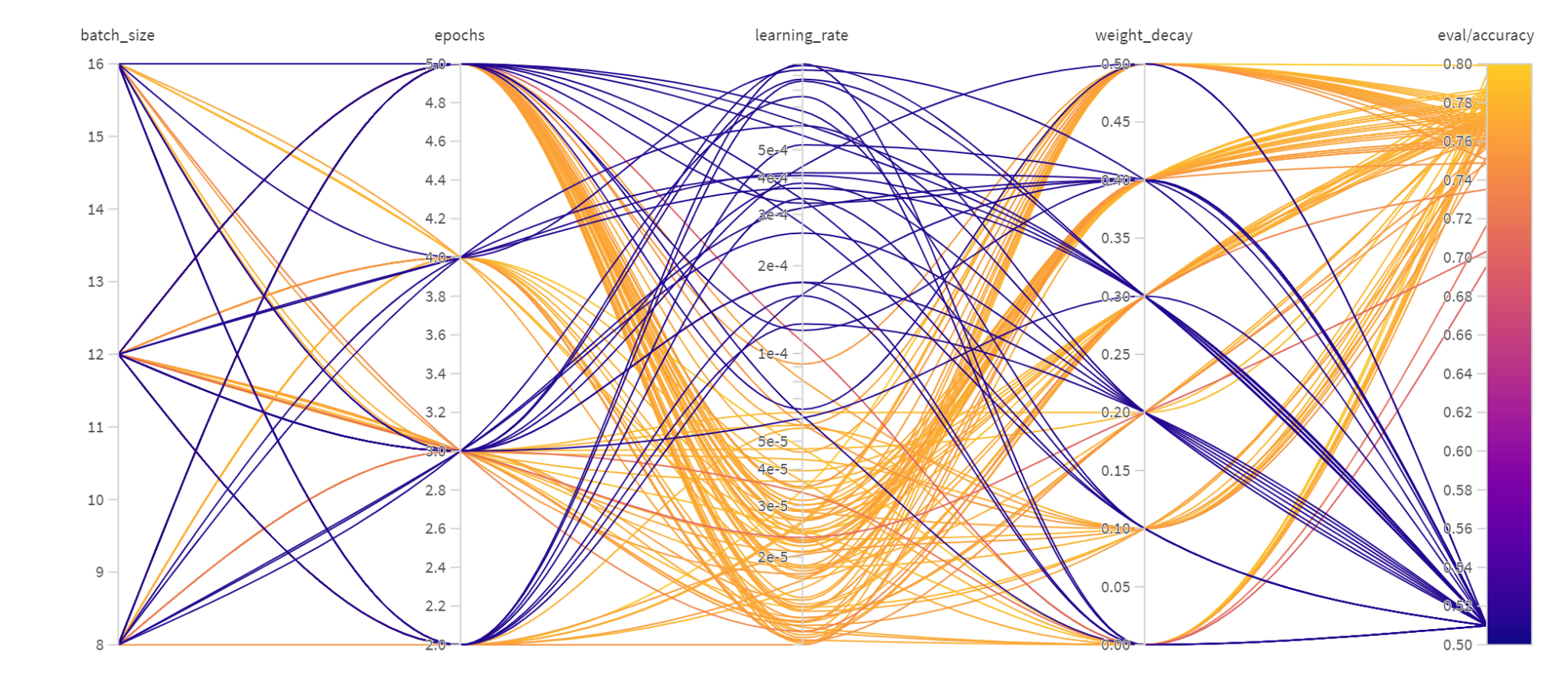}
    \caption{Hyperparameter Optimization runs for \texttt{multi} vs. \texttt{passage} and \texttt{phrase}}
    \label{fig:multi}
\end{figure*}

\begin{figure*}[t]
    \centering
    \includegraphics[width=15cm]{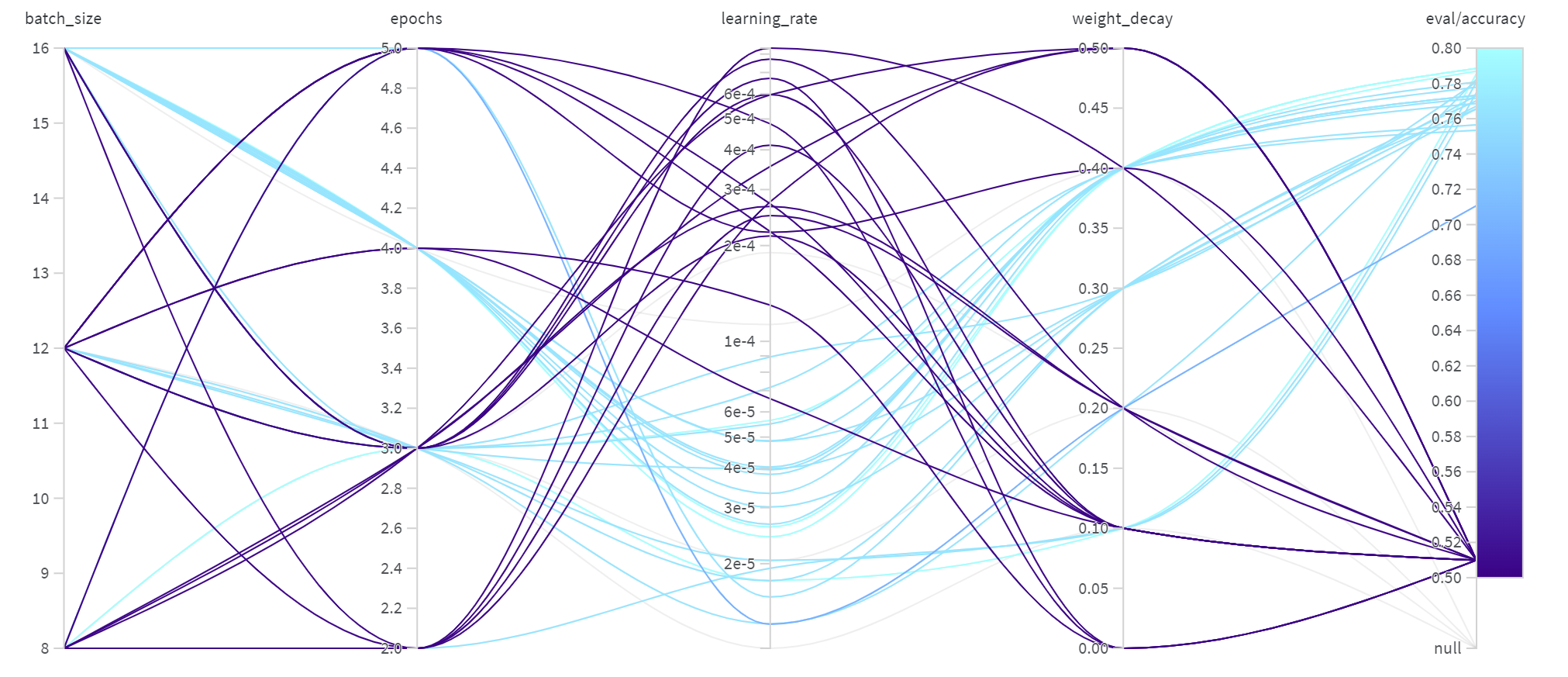}
    \caption{Hyperparameter Optimization runs for \texttt{passage} vs. \texttt{phrase}}   
    \label{fig:passagevphrase}
\end{figure*}

After defining the models, we utilized WandB's
hyperparameter search function to optimize our models. For both models, we sought to tune the batch size, epoch count, learning rate, and weight decay parameters. The batch size is the amount of samples that are passed to the model before it updates. The epoch count is the amount of iterations that the machine learning model goes through before the final model is trained. The learning rate is a parameter that determines how fast the model moves towards the optimal weigh values. The weight decay parameter regularizes the weights of the model to prevent overfitting. The search space that we used can be seen in Table \ref{tab:hyperparams_search}.
The results for Model 1 and 2 can be seen in Table \ref{tab:hyperparams_result}. The searches can be seen in Figure \ref{fig:multi} and Figure \ref{fig:passagevphrase}.
In the figures, each line is a "run" with certain parameters. The color of the line corresponds to the gradient on the right of the figure which represents the accuracy that the model had on its task. The parameter that had the highest correlation figure with accuracy was the learning rate. This is clear in the figure because most of the lines that have the color corresponding to the higher accuracy can be seen to have lower learning rates. 
\begin{table}
\centering
\begin{tabular}{lc}
\toprule
\textbf{Parameter} & \textbf{Search Space}\\
\midrule
Batch Size & [8,12,16] \\
Epoch & [2,3,4,5] \\
Learning Rate & $ 1 * 10^{-5} , 1* 10 ^{-3}$ \\ 
Weight Decay & [0.1, 0.2, 0.3, 0.4, 0.5] \\ 
\bottomrule
\end{tabular}
\caption{Search space for hyperparameter optimization for Task 1, clickbait classification.}
\label{tab:hyperparams_search}
\end{table}

\begin{table}
\centering
\begin{tabular}{lc}
\toprule
\textbf{Parameter} & \textbf{Value}\\
\midrule
Batch Size & 16 \\
Epoch & 4 \\
Learning Rate & $5.9 * 10^{-6}$ \\ 
Weight Decay & 0.4 \\ 
\bottomrule
\end{tabular}
\begin{tabular}{lc}
\toprule
\textbf{Parameter} & \textbf{Value}\\
\midrule
Batch Size & 16 \\
Epoch & 2 \\
Learning Rate & $ 8.2 * 10^{-6} $ \\ 
Weight Decay & 0.5 \\ 
\bottomrule
\end{tabular}
\caption{Optimal parameters for Model 1 and Model 2 respectively for Task 1, clickbait classification.}
\label{tab:hyperparams_result}
\end{table}

\subsection{Evaluation Metric}
We used balanced accuracy for the evaluation metric as the dataset did not have an equal amount of each spoiler. Balanced accuracy computes the mean of the sensitivity and specificity to prevent the imbalance of the dataset from affecting our performance. In our case, it would have affected model 1 because there was significantly more \verb|multi| than not. The equation for balanced accuracy is

\small{\[balanced-accuracy(y, \hat{y}, w) = \frac{1}{\sum{\hat{w}_i}} \sum_i 1(\hat{y}_i = y_i)\hat{w}_i  \]} \normalsize

where $y$ is the predicted value for the $i$th term, $\hat{y}$ is the reference value, and $\hat{w}$ is the sample weight adjusted with the function $\frac{w_i}{\sum_{j} 1(y_j = y_i)w_j}$. \cite{scikit-learn}
\subsection{Results}
For Model 1, using the optimal parameters in Table \ref{tab:hyperparams_result} gave a balanced accuracy of 0.7884. For Model 2, using the parameters above gave a balanced accuracy of 0.799. Thus, our models performed better than the existing baseline set by the creators of the dataset which had a balanced accuracy of 0.7157.
\section{Task 2: Clickbait Spoiling}
The second task that we tackle is the spoiling of the clickbait. We approach this task by using a Large Language Model (LLM) to generate the spoiler. Specifically, we utilize openAI's GPT-3 to output the spoiler. \cite{DBLP:journals/corr/abs-2005-14165}
\subsection{Prompt Engineering}
Prompt engineering is a key part of getting the full value from LLM models such as GPT-3. There are multiple ways to go about prompt engineering, but we used the one-shot method. This method includes giving the LLM one completed example in the prompt as an example that it can "learn" from. Since finetuning GPT-3 is expensive both computationally and financially due to model predictions being exclusive to the openAI API, approaches such as one-shot can help "train" the model to make the output better quality. We use three different prompts in the prediction process for the three different types of clickbait. The one-shot prompts contain a random entry from the train dataset with instruction on how the model should interpret the prompt. The prompts that we used can be found in Appendix \ref{prompts}.

\subsection{Evaluation Metric}
\label{evalmetric}
We chose the BLEU-4 metric to evaluate our outputs. \cite{papineni_roukos_ward_zhu_2001} BLEU is one of the metrics that the creators of the dataset chose to use in their evaluation of the baseline. This makes comparing our results very easy. BLEU is a score between 0 and 1 that calculates the similarity between two texts, a reference (gold) text and a predicted text. It computes the geometric average of the precision values for n-grams for n = 1 to n = 4 inclusive common to both texts, multiplied by a penalty value for short predicted texts. The equation for BLEU can be seen below where $\hat{S}$ is the input string, $S$ is the reference string, $BP$ represents the brevity penalty function defined as $e^{-(r/c -1)^+}$ where $c$ is the length of the input string and $r$ is the length of the reference string, $w_n$ is the weighing vector, and $p_n$ is the modified n-gram precision function defined as $p_n(\hat S; S):= \frac{\sum_{i=1}^M\sum_{s\in G_n( \hat y^{(i)})} \min(C(s, \hat y^{(i)}), \max_{y\in S_i} C(s, y))}{\sum_{i=1}^M\sum_{s\in G_n( \hat y^{(i)})} C(s, \hat y^{(i)})}$ which calculates the precision values without the brevity penalty. 

\small{\[BLEU _w(\hat{S}; S) := BP(\hat{S}; S) \cdot exp(\sum_{n =1}^\infty w_n \ln p_n (\hat{S}; S))\]} \normalsize
\
\subsection{Results}
The BLEU-4 score for task 2 was 0.13 on the validation set. This was significantly lower than the baseline put out by the dataset creators.
\section{Error Analysis}
\subsection{Task 1}
 In task 1, error could have come from either the input format or the hyperparameter optimization. The input format could have included the text of the article which might have led to a higher accuracy. The optimal set of hyperparameters might have been outside of search space, and therefore unreachable.
\subsection{Task 2}
In task 2, we obtained a score that was significantly lower than the baseline score. This performance could have been due to a variety of factors. First, the prompt could have been misleading to the LLM. The quality of the output depends on the quality of the prompt, and the instructions provided might have been too vague for the LLM to understand. However, the main problem with the score was that the evaluation metric was fundamentally flawed at evaluating the the similarity between generated text and a gold standard. The BLEU-4 method was one that directly compared the two strings. Thus, if the LLM generated its own spoiler that was not taken from the text of the article verbatim, it would be severely penalized. The original intent of the authors of the dataset was to approach the problem as an extractive question answering problem. Such an approach would result in higher BLEU scores because the model would only be able to output direct spans from the article text.
\section{Conclusion}
We developed solutions to two problems dealing with internet clickbait. First, we classified clickbait into three types: \verb|phrase|, \verb|passage|, and \verb|multi|. Next, we spoiled the clickbait with a short summary of the article. The output of the first task helped with the second task as it allowed us to use a specific prompt on the LLM for the specific type of clickbait. The first task used a basic sequence classification NLP model while the second task utilized GPT-3 and prompt engineering to output a spoiler. The first task resulted in balanced accuracy score of 0.7884 and 0.799 which outperformed the baseline scores significantly.  The second task resulted a BLEU score of 0.13 which was lower than the baseline model; however, this was due to the evaluation metric being a poor fit for generative NLP models. 
\section{Future Work}
\subsection{Research}
\subsubsection{Task 1}
For task 1, a broader hyperparameter search might be result in better accuracy values. Since we were limited by our local hardware of an RTX 3070 graphics card, we could only use batch size values with a max of 24 before we encountered out-of-memory (OOM) errors. We could also experiment with different input formats to try and increase performance.
\subsubsection{Task 2}
For task 2, experimenting with a few-shot approach instead of a one-shot approach might have increased performance. We used one-shot because any other method resulted in the input being too large for the API to handle. Furthermore, another evaluation metric that is more accepting of generative text could help boost performance.
\subsection{Applications}
A major application of this work is in internet browsers. Creating an extension to spoil clickbait before users click on clickbait titles would help solves many issues related to the massive proliferation of disinformation happening on the Internet by quenching users' demand for the information that the clickbait title lures them with.

\nocite{*}
\bibliographystyle{abbrv}
\bibliography{bibliography}
\appendix
\section{Appendix: Prompts}
\label{prompts}
\subsection{Multi}
Predict the spoiler to the clickbait title similar to the example below:

\verb|Title|: Six lessons from the godfather of California cuisine \\

\verb|Article|: [...]The Huffington Post recently caught up with Patterson to see if we could glean any morsels of wisdom from the self-taught chef and father of two. Herewith, a few lessons from an icon most certainly qualified to discuss the state of California cuisine and beyond. Patterson, right, preparing food in Madrid. 1) Eat your veggies.[...]\footnote{Truncated for space.} 
\\

\verb|Spoiler|: Daniel Patterson, 1) Eat your veggies., 2) Enjoy said veggies a few weeks after their season starts.", 3) Ingredients dictate everything., 4) Don't serve complex foods to your 4-year-old.", 5) You can succeed without a mentor. 
\subsection{Passage}
Predict the spoiler to the clickbait title similar to the example below:

\verb|Title|: Instagram Just Killed This Feature\\

\verb|Article|: After adding a much requested pinch and zoom feature last week, Facebook-owned photo sharing giant Instagram is taking away a popular feature-photo maps.According to technology news site Mashable, Instagram is killing the feature that allowed you to see a map of where a given user's photos were taken (if they designated a location for the photo). You could also zoom in on maps to get a closer view of where photos were taken in a particular city or region.[...]\footnote{Truncated for space.}\\

\verb|Spoiler|: the feature that allowed you to see a map of where a given user's photos were taken (if they designated a location for the photo). 
\subsection{Phrase}
Predict the spoiler to the clickbait title similar to the example below:

\verb|Title|: The cheapest place for a last-minute half-term holiday\\

\verb|Article|: Cyprus is the cheapest option for a last-minute February half-term getaway, new research has shown. Paphos, on the island's west coast, and also one of 2017’s two European Capitals of Culture, ranked first in an analysis of package holiday prices and on-the-ground costs in 10 popular winter sun destinations.[...]\footnote{Truncated for space.}\\
\verb|Spoiler|: Cyprus 

\end{document}